\documentclass[10pt,twocolumn,letterpaper]{article}

\usepackage{iccv}              

\newcommand{\czx}[1]{#1}

\newcommand{\methodname}{{NAVER}}
\newcommand{\dfa}{{DFA}}

\usepackage{pifont}
\usepackage{float}

\usepackage{tcolorbox}
\usepackage{url}
\usepackage{xcolor}
\definecolor{newcolor}{rgb}{.8,.349,.1}
\usepackage{enumitem}
\usepackage{colortbl}
\usepackage{float}
\usepackage[ruled]{algorithm2e}
\usepackage{subcaption}
\usepackage{multirow}
\usepackage{enumitem}
\usepackage[T1]{fontenc}
\tcbuselibrary{breakable}

\definecolor{mygray}{gray}{.9}
\newtcolorbox[list inside=prompt,auto counter,number within=section]{prompt}[1][]{
    colbacktitle=black!60,
    coltitle=white,
    fontupper=\footnotesize,
    boxsep=5pt,
    left=0pt,
    right=0pt,
    top=0pt,
    bottom=0pt,
    boxrule=1pt,
    title={#1},
    #1, 
}

\definecolor{iccvblue}{rgb}{0.21,0.49,0.74}
\usepackage[pagebackref,breaklinks,colorlinks,allcolors=iccvblue]{hyperref}

\def\paperID{6765} 
\def\confName{ICCV}
\def\confYear{2025}

\title{\includegraphics[height=20pt,width=20pt,trim=0 7mm 0 -7mm]{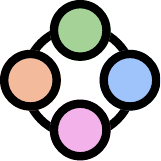} \methodname{}: A Neuro-Symbolic Compositional Automaton for Visual Grounding with Explicit Logic Reasoning}

\author{Zhixi Cai, Fucai Ke, Simindokht Jahangard, Maria Garcia de la Banda, \\ Reza Haffari, Peter J. Stuckey, Hamid Rezatofighi\\
Monash University\\
{\tt\small \{zhixi.cai,fucai.ke1,simindokht.jahangard,maria.garciadelabanda,}\\
{\tt\small gholamreza.haffari,peter.stuckey,hamid.rezatofighi\}@monash.edu}
}

\begin{document}

\twocolumn[{
\maketitle
\begin{center}
\captionsetup{type=figure}
\includegraphics[width=\textwidth]{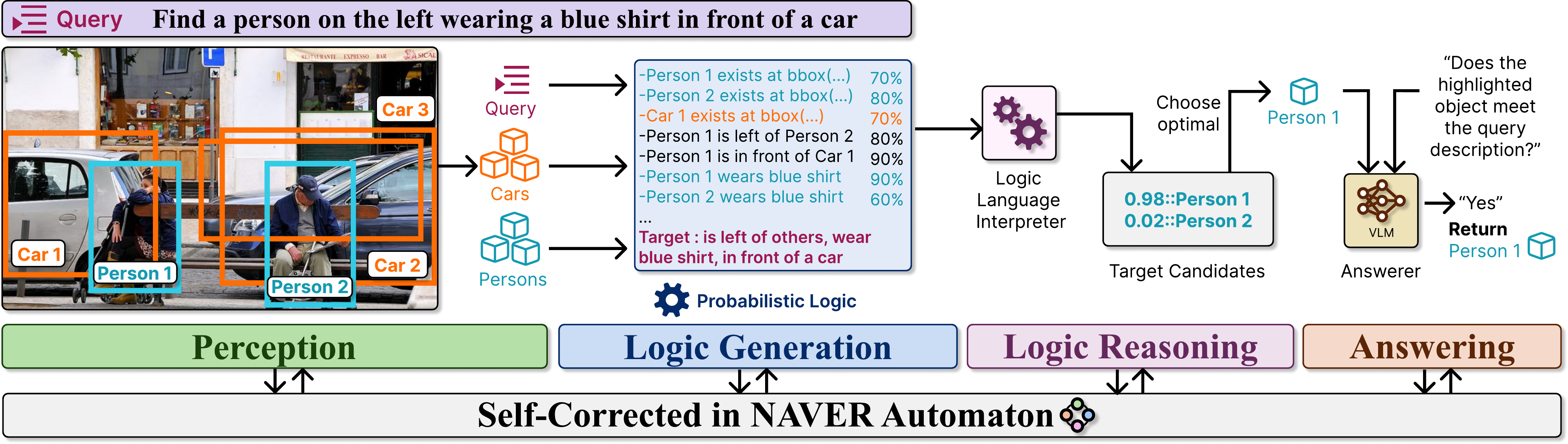} 
\vspace{-6.5mm}
\captionof{figure}{
\textbf{Overview of \methodname{}}. Given the query ``A person on the left wearing a blue shirt in front of a car'', \methodname{} first identifies the entities of interest, their relations (\eg ``Person 1 is left of Person 2'', ``Person 1 is in front of Car 1''), and attributes (\eg ``Person 1 is wearing blue shirt'').  It then transforms these statements into ProbLog~\cite{de_raedt_problog_2007} logic expressions, and applies probabilistic reasoning to find the correct match. The validated output identifies ``Person 1'' as the correct result, demonstrating \methodname{}'s ability for complex VG tasks.
}
\label{fig:teaser}
\end{center}
}]

\begin{abstract}
Visual Grounding (VG) tasks, such as referring expression detection and segmentation tasks are important for linking visual entities to context, especially in complex reasoning tasks that require detailed query interpretation. This paper explores VG beyond basic perception, highlighting challenges for methods that require reasoning like human cognition. Recent advances in large language methods (LLMs) and Vision-Language methods (VLMs) have improved abilities for visual comprehension, contextual understanding, and reasoning. These methods are mainly split into end-to-end and compositional methods, with the latter offering more flexibility. Compositional approaches that integrate LLMs and foundation models show promising performance but still struggle with complex reasoning with language-based logical representations. To address these limitations, we propose \methodname{}, a compositional visual grounding method that integrates explicit probabilistic logic reasoning within a finite-state automaton, equipped with a self-correcting mechanism. This design improves robustness and interpretability in inference through explicit logic reasoning. Our results show that \methodname{} achieves SoTA performance compared to recent end-to-end and compositional baselines. 
The code is available at \url{https://github.com/ControlNet/NAVER}.
\end{abstract}    
\vspace{-5mm}
\section{Introduction}
Grounding referring expressions in image tasks, including detection and segmentation~\cite{kazemzadeh_referitgame_2014}, are fundamental to visual reasoning and cover a wide range of complexities~\cite{yang_cross-modal_2019}.
This paper explores a level of visual grounding that emphasizes reasoning and goes beyond basic perception~\cite{yu_modeling_2016}. 
This is challenging for current models because it requires interpreting complex queries, creating layered visual representations, and performing step-by-step reasoning for coherent interpretations and contextually appropriate inferences, closely resembling human cognitive functions~\cite{amizadeh_neuro-symbolic_2020}.
For example (See~\autoref{fig:teaser}), the query ``find a person on the left wearing a blue shirt in front of a car'' requires the method to process multiple components simultaneously. This includes detecting entities like ``person'', ``car'' and others, recognizing attributes such as ``wearing blue shirt'' and identifying the relations like ``on the left'' and ``in front of'' between these entities to generate a contextually grounded understanding of the visual content. 

Recent advances in large language models (LLMs)~\cite{chowdhery_palm_2023,bai_constitutional_2022,ouyang_training_2022,radford_language_2019} and their derivatives,
Vision-Language Models (VLMs)~\cite{li_otter_2023,wu_next-gpt_2023,stanic_towards_2024,zhu_minigpt-4_2023} have increased expectations for better visual grounding through enhanced reasoning. These developments can be categorized into two main types: (1) end-to-end (monolithic) methods that integrate various tasks into a unified model, and (2) compositional methods that leverage modular components to improve flexibility and interpretability in inference.
End-to-end methods demonstrate promising results, but often struggle with complex reasoning tasks such as geometry relations~\cite{chen_unigeo_2022}, and their reasoning capabilities are implicit and thus difficult to verify. The challenge may be due to missing explicit reasoning structures, or over-reliance on end-to-end learning, or the difficulty in disentangling reasoning components.
In contrast, compositional approaches are more effective in reasoning-based visual grounding challenges, as they decompose complex tasks into simpler components using a divide-and-conquer strategy.
They integrate LLMs as planners, code generators, or reasoners, with Vision Foundation Models (VFMs) or Vision Language Model (VLM) for visual perception, enabling structured analysis and task-specific planning to enhance adaptability and generalization across scenarios ~\cite{gupta_visual_2023, suris_vipergpt_2023, lu_chameleon_2023, wu_visual_2023}. Thus, the reasoning of these models is interpretable and verifiable, providing a clear framework for decision-making.
However, current compositional methods have several limitations.
First, they primarily rely on the commonsense knowledge encoded in LLMs for planning and reasoning, which may not fully capture the complex relations, \eg physics-based and geometric relations, needed for comprehensive understanding in complex queries~\cite{chen_spatialvlm_2024}.
Second, logic is represented in natural language, which lacks the necessary structure for effective reasoning, resulting in ambiguous interpretations and difficulties in extracting relevant information, limiting their ability to perform accurate and context-aware reasoning. 
Finally, if any mistakes or failures occur at any step in the process, they may accumulate throughout inference, as these systems lack an automatic self-correction mechanism.

In this paper, we introduce a novel compositional visual grounding method named \emph{Neuro-symbolic compositional Automaton for Visual grounding with Explicit logic Reasoning} (\methodname{}) that addresses the limitation of existing compositional frameworks by \emph{(i)} integrating explicit probabilistic logic reasoning using ProbLog~\cite{de_raedt_problog_2007} within a compositional pipeline to enhance visual reasoning capabilities, and \emph{(ii)} incorporating a flexible, deterministic finite-state automaton (\dfa)\footnote{For simplicity, ``automaton'' refers to ``deterministic finite-state automaton'' in this paper.} with a self-correction mechanism.
Our method combines symbolic logic with large language models (LLMs) and neural-based vision foundation models, providing an explainable and logic inference for complex queries. By integrating symbolic reasoning within a \dfa-based pipeline, \methodname{} dynamically navigates between states based on intermediate results and specific conditions, handling complex contextual constraints and relations in reasoning-based grounding. In addition, \methodname{} has been designed so that each step and module includes a self-correction mechanism, ensuring that mistakes occurring in one step or module, do not affect others. This combination not only improves the system's ability to handle complex queries but also provides transparency and interpretability, thus increasing trust and facilitating error analysis. We conducted extensive experiments on popular datasets, showing that \methodname{} achieves state-of-the-art (SoTA) performance, comparing with SoTA end-to-end~\cite{li_grounded_2022, liu_grounding_2023, peng_kosmos-2_2023, dai_simvg_2024, xiao_florence-2_2024} and compositional~\cite{stanic_towards_2024, suris_vipergpt_2023, ke_hydra_2024} methods. These results are supported by detailed ablation studies validating the effectiveness of each component.
In summary, the key contributions of this work are as follows: 
\begin{enumerate} 
    \item We propose \methodname{}, a novel compositional visual grounding method that leverages explicit logic representation for complex reasoning tasks.
    \item We design a flexible, deterministic finite-state automaton-based system that dynamically transitions between states based on intermediate results, incorporating a self-correction mechanism at each step to improve robustness and adaptability.
    \item We perform comprehensive experiments on standard benchmarks, demonstrating that \methodname{} achieves state-of-the-art performance, with additional ablation studies to validate the effectiveness of each component. 
\end{enumerate}

\begin{figure*}[t]
\centering
\includegraphics[width=\textwidth]{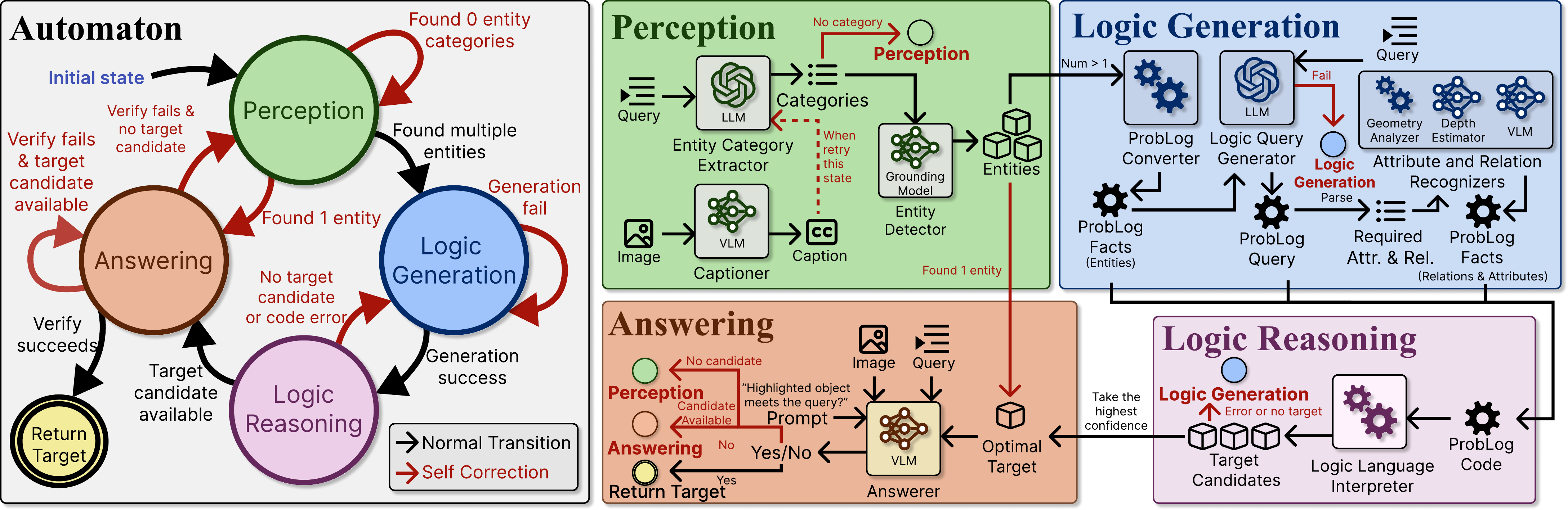}
\vspace{-6.5mm}
\caption{\textbf{Pipeline of \methodname{}.} \methodname{} is organized as a \emph{deterministic finite-state automaton} (\dfa) with five states: (1) \emph{Perception}, which is the initial state where relevant visual information is extracted by identifying entity categories and localizing entities; (2) \emph{Logic Generation}, where an LLM uses the caption and query to generate logic expressions in ProbLog, incorporating entities, and ProbLog query. The relations and attributes required by the query are recognized by the foundation models and converted to  ProbLog code; (3) \emph{Logic Reasoning}, where ProbLog code is executed to identify the target candidates based on explicit probabilistic logic; (4) \emph{Answering}, where the identified optimal target candidate is validated by a VLM to check if it fits the query; and (5) \emph{Return Target}, which is the final state only reached if the check succeeds. The \dfa\ structure (left) coordinates the flow across these stages, with conditional transitions that enable a self-correcting mechanism (\textcolor{red}{red arrows}) to revisit previous states if refinement is needed, ensuring the robustness and interpretability in the final response. Note that the automaton is deterministic, as the inputs that govern all transitions departing each state are mutually exclusive. 
}
\label{fig:pipeline}
\vspace{-4.5mm}
\end{figure*}

\section{Related Works}
\label{sec:related_works}

Recent advances in LLMs~\cite{chowdhery_palm_2023,bai_constitutional_2022,ouyang_training_2022,radford_language_2019} and VLMs~\cite{li_otter_2023,wu_next-gpt_2023,stanic_towards_2024,zhu_minigpt-4_2023} have raised expectations for improved visual grounding in Referring Expression Detection~\cite{kazemzadeh_referitgame_2014} and Referring Expression Segmentation~\cite{liu_gres_2023} tasks. The Referring Expression Detection involves identifying objects in images based on natural language descriptions, while Referring Expression Segmentation goes further by providing precise pixel-level segmentation of those objects. These approaches can be categorized into two main streams: \emph{end-to-end} and \emph{compositional methods}.

\noindent\textbf{End-to-End Methods.}
End-to-end methods in visual grounding aim to leverage neural networks to directly map text queries to objects in an image, without explicitly separating the stages of grounding and reasoning.
Early models mainly use transformer-based architectures trained on extensive datasets to achieve generalized language and vision understanding~\cite{heigold_video_2023, minderer_scaling_2023, li_grounded_2022, liu_grounding_2023, cheng_yolo-world_2024}.  With the recent advances in LLMs~\cite{openai_gpt-4o_2024, dubeyLlama2024, brown_language_2020} and VLMs~\cite{li_otter_2023, alayrac_flamingo_2022, dai_instructblip_2023, wang_visionllm_2023, wu_next-gpt_2023, liu_visual_2023, chen_minigpt-v2_2023}, these architectures have further enhanced performance in visual grounding tasks~\cite{peng_kosmos-2_2023, lai_lisa_2024, rasheed_glamm_2024, pi_perceptiongpt_2024, xia_gsva_2024}.
However, end-to-end methods still face limitations, including low interpretability and high computational cost due to the demands of training neural networks on large-scale datasets. Their data-driven nature often limits their capacity for generalization and analytical tasks, particularly those involving geometry and relational reasoning, which require a more context-aware approach~\cite{chen_unigeo_2022, villalobos_will_2024, gupta_visual_2023}.

\noindent\textbf{Compositional Visual Reasoning Methods.}
The compositional strategy offers a solution to the difficulties encountered by end-to-end methods~\cite{suris_vipergpt_2023,you_idealgpt_2023,lu_chameleon_2023,gupta_visual_2023,stanic_towards_2024}. Compositional methods address complex problems by dividing them into subtasks, solving each task independently, and then using intermediate results to solve the initial problem. By utilizing the powerful chain-of-thought (CoT) capabilities of LLMs, these methods can effectively simplify complicated challenges into smaller, solvable steps through specific instructions~\cite{gupta_visual_2023,suris_vipergpt_2023}. These instructions may include execution code that includes logical operations. 
VisProg~\cite{gupta_visual_2023} and ViperGPT~\cite{suris_vipergpt_2023} reduce task-specific training by using code generation LLMs, integrating VLMs into the program to generate final output. HYDRA~\cite{ke_hydra_2024} combines a planner, RL agent, and reasoner for reasoning, outperforming other methods. It uses predefined instructions to activate perception tools and combine data for reasoning.
Despite their strengths, these methods have notable limitations. They heavily rely on LLMs for reasoning, which may not fully capture complex relations like physics or geometry. Additionally, logic representations in natural language lead to ambiguity, making it hard to extract relevant information. Lastly, these systems lack self-correction mechanisms, allowing errors to propagate through the inference steps. Addressing these issues with formulated explicit logic representations and robust reasoning could enhance performance and interpretability.
To address these issues, we propose \methodname{}, a method that improves visual reasoning by integrating symbolic logic, probabilistic logic, and a self-correcting deterministic finite-state automaton. A more detailed description of our method is provided in \autoref{sec:method}.
\section{\methodname{}}
\label{sec:method}

\methodname{} addresses complex visual grounding tasks that require complex reasoning, building on the limitations of previous works~\cite{gupta_visual_2023, suris_vipergpt_2023, lu_chameleon_2023, ke_hydra_2024}. In this work, we focus on referring expression detection and segmentation as the VG tasks that require reasoning with object attributes and relations among two or multiple entities within an image. Given the input $X=\{I, Q\}$ including image $I$ and a textual query $Q$, the objective of \methodname{} is to produce reliable and interpretable grounding result $Y$ satisfying the query $Q$. The proposed method is formulated as a \emph{deterministic finite-state automaton} for a flexible pipeline, and an explicit logic reasoning ability using ProbLog~\cite{de_raedt_problog_2007} code generated by an LLM. To improve the robustness of the system, the full-chain of the pipeline is validated in the automaton. With this design, the system maintains a similar computational complexity to that of previous compositional methods~\cite{ke_hydra_2024}, but achieves better performance (See \autoref{sec:quantitative}). An overview of the framework is shown in \autoref{fig:pipeline}.

\subsection{Deterministic Finite-State Automaton}
\label{sec:automaton}

A \emph{Deterministic Finite-State Automaton} (\dfa) is a mathematical model used to describe a system that moves deterministically between a finite number of defined states in response to specific inputs. Formally, a \dfa\ is a 5-tuple $\langle S, \Sigma, \delta, s_o, F \rangle$, where $S$ is a finite set of states, $\Sigma$ is a finite set of symbols, $\delta: S \times \Sigma \mapsto S$ a transition function between states based on the input symbol, $s_0 \in S$ an initial state, and $F\subseteq S$ a set of final states. Alternatively, a \dfa\ can be represented by a directed graph (or state diagram) whose vertices represent the states in $S$; its edges are labeled with elements from $\Sigma$ representing the transitions from $\delta$; $s_0$ is the only vertex with a single, empty incoming edge; and the final states in $F$ are indicated by double circles.

The design of \methodname{} is based on the \dfa\ shown in~\autoref{fig:pipeline}, which has five states: \emph{Perception} ($S_P$) which is the initial state, \emph{Logic Generation} ($S_L$), \emph{Logic Reasoning} ($S_R$),  \emph{Answering} ($S_A$), and \emph{Return Target} ($S_F$) which is the final state. The \dfa\ starts in the initial state $S_P$ and moves across states deterministically based on the transition function represented in \autoref{tab:transitions}, where the result of each condition maps to an element of alphabet $\Sigma$. Note that the \dfa\ can return to a previously visited state based on self-correction, thus improving the robustness of the pipeline. To prevent infinite loops within the \dfa, the system is programmed to terminate (i.e., to reach the final state \emph{Return Target}) after a defined number of retries during self-correction transitions (red arrow in \autoref{fig:pipeline}). Upon termination, the system returns the grounding target based on the currently collected information. 

Importantly, and unlike the conventional sequential pipelines used in previous compositional methods~\cite{ke_hydra_2024, shen_hugginggpt_2023, suris_vipergpt_2023, you_idealgpt_2023} where each step strictly follows the previous one, a \dfa{} allows for dynamic transitions between states based on intermediate results and specific conditions. This flexibility is critical not only for the self-correction mechanism mentioned above, but also for handling complex reasoning tasks, where the requirements for accurately visual grounding may vary depending on the context. 
The next subsections discuss the conditions and transitions of the \dfa{}.


\begin{table}[t]
\centering
\scalebox{0.75}{
\begin{tabular}{c|ccc}
\toprule[0.4mm]
\rowcolor{mygray} \textbf{\#} & \textbf{Current} & \textbf{Condition} & \textbf{Next} \\ \hline \hline
1 & $S_P$ & Number of entities categories $|C|=0$. & $S_P$ \\
2 & $S_P$ & Number of entities $|E|<2$. & $S_A$ \\
3 & $S_P$ & Number of entities $|E|\geq 2$. & $S_L$ \\ \hline
4 & $S_L$ & ProbLog code generation error & $S_L$ \\
5 & $S_L$ & ProbLog code generation success & $S_R$ \\ \hline
6 & $S_R$ & Number of result $|Y_L|=0$ & $S_L$ \\
7 & $S_R$ & Number of result $|Y_L|>0$ & $S_A$ \\ \hline
8 & $S_A$ & Answer ``Yes'' & $S_F$ \\
9 & $S_A$ & ``No'' and number of alternative results $|Y_L'|>0$ & $S_A$ \\
10 & $S_A$ & ``No'' and number of alternative results $|Y_L'|=0$ & $S_P$ \\
\bottomrule[0.4mm]
\end{tabular}}
\vspace{-2.5mm}
\caption{\textbf{State transition table for \methodname{} finite-state automaton.} Please refer to the \autoref{sec:perception} to
\autoref{sec:answering} for the details of transitions.}
\label{tab:transitions}
\vspace{-4.5mm}
\end{table}

\subsection{Perception (\texorpdfstring{$\boldsymbol{S_P}$}{S\_P})}
\label{sec:perception}


The \emph{Perception} state is responsible for establishing an initial understanding of the image content in relation to the query, serving as the basis for explicit reasoning in subsequent states. This state is designed to extract and encode relevant visual entities. By extracting these elements, $S_P$ creates a filtered representation of the image, focusing only on entities relevant to the query. The motivation for designing a dedicated \emph{Perception} state is to prepare the clean and formulated entities information for the following \emph{Logic Generation} state, making the system more interpretable and reducing computational complexity for later states.

For establishing a semantic understanding between the query and the image, and helping later states focus on the most relevant entities, the \emph{Perception} state begins by using a Vision-Language Model (VLM) as captioner to generate a caption $I_c$ for the image $I$. Given the textual query $Q$, an LLM identifies and extracts entity categories of interest $C = \{C_1, C_2, ...\}$ by analyzing both the caption $I_c$ and the query $Q$. The $I_c$ will be shown to the LLM when using only the query $Q$ is not enough for finding the target $Y$. This step filters out irrelevant objects, highlighting only those that are likely to contribute to the reasoning and localize the target.

Once the relevant entity categories $C$ are extracted, a foundation grounding model is applied as an Entity Detector to localize entities $E=\{E_1, E_2, ...\}$ belonging to the categories $C$ within the image, obtaining precise bounding boxes or segmentation masks $Y_E$. Entity IDs are auto-generated to the entities for building the mapping between the location information $Y_E$ and logic symbols. The probability $P(E)$ is assigned by the confidences of the grounding model detection.

\textbf{State Transition:} If no entity categories of interest are detected ($|C| = 0$), the automaton remains in $S_P$ to retry perception with feedback prompting to LLM (Transition 1 in \autoref{tab:transitions}). If no entity is detected ($|E| = 0$), the fallback result detected by the entity detector with query $Q$ is used as the sub-optimal solution. If only one entity is detected ($|E| = 1$), no logic reasoning is needed as the entity can be the final output result, and the automaton directly transitions to Answering $S_A$ (Transition 2) for final verification. If two or more entities are detected ($|E| \geq 2$), the automaton moves to Logic Generation ($S_L$) to proceed with logic reasoning (Transition 3).

\subsection{Logic Generation (\texorpdfstring{$\boldsymbol{S_L}$}{S\_L})}
\label{sec:logic_generation}

To translate the extracted visual information and the text query into explicit logic representations for reasoning, the \emph{Logic Generation} state is designed to generate the ProbLog~\cite{de_raedt_problog_2007} logic code for reasoning. ProbLog is selected as the logic representation because it supports probabilistic logic, allowing the system to reason about the uncertainties that naturally arise in visual tasks (e.g., overlapping entities or ambiguous attributes). 

To model and reason about the visual content and query in logic, the process in the \emph{Logic Generation} state begins by taking the outputs from the \emph{Perception} state, which include the identified entities $E$ with their spatial locations $Y_E$,  grounding confidences $P(E)$ and associated categories $C$. Leveraging predefined ProbLog code templates, each entity is converted into a ProbLog \emph{fact} that declares its ID, category, bounding box coordinates, and probability. 
Code \hyperref[code:problog]{3.1} gives an example.

To translate the natural language query $Q$ into a ProbLog \emph{rule}, an LLM is used to interpret $Q$ in the context of the already encoded entities and relations. In addition, the relations and attributes of entities are involved to model the query for better abstraction and generalization abilities. The LLM receives as input the encoded ProbLog facts, along with the original query $Q$, and generates the corresponding ProbLog rule for querying the existence of target $Y$. For example, if the query is ``Find the person on the left wearing a blue shirt'', the LLM produces a ProbLog rule that matches any entity fulfilling these conditions from the query $Q$. 

However, while the relations and attributes may be included in the target ProbLog rule, they have not yet been perceived. To handle this, we parse the ProbLog query generated by the LLM and extract any relation symbols $\Gamma_R=\{\Gamma_{R1}, \Gamma_{R2}, ...\}$ and attribute symbols $\Gamma_A = \{\Gamma_{A1}, \Gamma_{A2}, ...\}$. Given the location of entity $Y_E$ and the potential attributes $\Gamma_A$, an attribute recognizer VLM is prompted to produce the probability of each attribute $P(A)$ being present in each entity of interest $E$. For example, if the query requires identifying ``Find the person on the left wearing blue shirt,'' the VLM will assess each entity's likelihood of having the attribute ``wearing blue shirt''. The attribute probabilities are then converted into ProbLog facts as before. Similarly, a relation recognizer including a VLM, a symbolic geometry analyzer and a depth estimator, is employed to produce the single-directional relations $R = \{R_{1,2}, R_{1,3}, ...\}$ with the probabilities $P(R)$. See Code \hyperref[code:problog]{3.1} for an example of ProbLog generated code.

\textbf{State Transition:} If an error is encountered during ProbLog code generation, the automaton stays in $S_L$ and retries the generation process with LLM (Transition 4 in \autoref{tab:transitions}). Upon successful generation of ProbLog code, the automaton transitions to the \emph{Logic Reasoning} state $S_R$ to proceed probabilistic inference (Transition 5).

\subsection{Logic Reasoning (\texorpdfstring{$\boldsymbol{S_R}$}{S\_R})}
\label{sec:logic_reasoning}

The \emph{Logic Reasoning} state performs probabilistic inference to identify the entities $E$ that best satisfies the query conditions. Probabilistic reasoning is essential here because it allows the system to weigh potential matches based on confidence levels when handling visual ambiguity. Using the logic programming language interpreter, the logic code representing entities, relations, attributes and query is executed to produce the target candidates $Y_L = \{Y_{L1}, Y_{L2}, ...\}$, with each result ranked according to its probability $P(Y_L)$. 
The candidates produced in this step is the input for final confirmation in the \emph{Answering} state, providing a robust filtered list that can be verified with high confidence.

\textbf{State Transition:} If no results are produced ($|Y_L| = 0$), the automaton returns to the \emph{Logic Generation} state ($S_L$) to refine the logic expressions generated for the query (Transition 6 in \autoref{tab:transitions}). If one or more results are identified ($|Y_L| > 0$), the automaton proceeds to the \emph{Answering} state ($S_A$) for final verification (Transition 7).

\begin{prompt}[title={Code \thetcbcounter: ProbLog Code for Explicit Logic}] \label{code:problog}
\% entity(ID: str, category: str, x1: int, y1: int, x2: int, y2: int).\\
0.7435::entity(``person\_0'', ``person'', 360, 171, 480, 386).\\
0.4134::entity(``person\_1'', ``person'', 0, 142, 159, 478).\\
...\\
\% relation(entity\_a: str, entity\_b: str, value: str).\\
0.0001::relation(``person\_0'', ``person\_1'', ``is'').\\
0.9986::relation(``person\_0'', ``person\_1'', ``left of'').\\
\% attribute(entity: str, value: str).\\
0.8711::attribute(``person\_0'', ``wearing\_blue\_shirt'').\\
0.1468::attribute(``person\_1'', ``wearing\_blue\_shirt'').\\
...\\
\% find person on the left wearing blue shirt\\
target(ID) :- entity(ID, ``person'', \_, \_, \_, \_), relation(ID, \_, ``left of''), attribute(ID, ``wearing\_blue\_shirt'').\\
query(target(ID)).
\end{prompt}

\subsection{Answering (\texorpdfstring{$\boldsymbol{S_A}$}{S\_A})}
\label{sec:answering}

The \emph{Answering} state serves as a final verification step to ensure that the top-ranked candidate from the \emph{Logic Reasoning} state aligns with the query’s requirements. The reasoning output is prompted to an answerer VLM to confirm if the predicted target entity meets the query. This validation step ensures the output aligns with the query’s intent before returning it as the final result.

To help the answerer VLM better access the top-ranked result $V_{L1}$ from the \emph{Logic Reasoning} state in the image $I$, the image is annotated with $V_{L1}$ to highlight the identified entity within the visual context~\cite{yang_set--mark_2023}. This annotated image is given as input to the answerer VLM, with a prompt asking whether the annotated object satisfies the query $Q$. To ensure a controlled response, we use the next-token prediction of VLM to calculate the probability of receiving a ``Yes'' or ``No'' answer, and the response with the higher probability is selected.

\textbf{State Transition:} If the answerer responds with ``Yes”, the automaton proceeds to the final state $S_F$ and returns the $Y_{L1}$ as the \textbf{final output} (Transition 8 in \autoref{tab:transitions}). If the response is ``No” and there are alternative results ($|Y_L'| > 0$) where $Y_L' = Y_L \setminus \{Y_{L1}\}$, the automaton stays in $S_A$ to evaluate the top-ranked candidate of the remaining candidate results $Y_L := Y_L'$ (Transition 9). If the response is “No” and there are no other candidates ($|Y_L'| = 0$), the automaton returns to Perception $S_P$ to gather new visual information (Transition 10). 

\section{Experiments and Results}

\subsection{Implementation Details}

We implement the method using PyTorch~\cite{paszke_pytorch_2019} with Nvidia L40s GPUs. The vision foundation models and LLMs used in \methodname{} are flexible and can be exchanged for any other methods of the same type. Specifically, we use GPT-4o Mini~\cite{openai_gpt-4o_2024} for all LLMs in the system and other compositional methods~\cite{ke_hydra_2024, suris_vipergpt_2023}. We use InternVL2-8B~\cite{chen_internvl_2024} as the VLM captioner in the \emph{Perception} state and as answerer in the \emph{Answering} state, while XVLM~\cite{zeng_multi-grained_2022} is used as attribute recognizer in the \emph{Logic Generation} state. We also use DepthAnythingV2~\cite{yang_depth_2024} as depth estimator and Florence2-L~\cite{liu_grounding_2023} as entity detector. For converting the bounding boxes to segmentation masks, SAM~\cite{kirillov_segment_2023} is used.
For segmentation tasks, we utilize GLaMM~\cite{rasheed_glamm_2024} for SoTA performance. For fair comparison with the previous compositional baselines~\cite{ke_hydra_2024, suris_vipergpt_2023}, we evaluated two versions of \methodname{}: (1) using the same foundation models (including GLIP~\cite{li_grounded_2022}) as HYDRA and ViperGPT, (2) using  the SoTA foundation models for best performance. We discuss the choice and the performance analysis of using different LLMs and grounding methods in \autoref{sec:quantitative} and \autoref{sec:ablation}. 
For probabilistic logic reasoning, we used Scallop~\cite{li_scallop_2023} to enhance efficiency. The initial logic code is generated in ProbLog, then automatically translated into Scallop, a Rust-based probabilistic logic programming language optimized for high-performance inference. This translation maintains all probabilistic logic expressions while enabling faster reasoning execution than traditional ProbLog implementations, especially when dealing with large logic programs. 
We implement the previous baselines~\cite{ke_hydra_2024, suris_vipergpt_2023, li_grounded_2022, liu_grounding_2023, cheng_yolo-world_2024, kirillov_segment_2023, rasheed_glamm_2024, dai_simvg_2024, xiao_florence-2_2024} using the official source code, and borrow the results of other baselines from \cite{ke_hydra_2024, lai_lisa_2024, zhang_psalm_2024}. More details of the implementations will be elaborated in the supplementary material. 

\subsection{Task, Dataset and Metric}
This work focuses on referring expression detection and segmentation as the two visual grounding tasks that require reasoning with object attributes and relations among multiple elements within an image.  For detection, we evaluate on RefCOCO test A, RefCOCO+ test A, RefCOCOg~\cite{kazemzadeh_referitgame_2014} test and Ref-Adv~\cite{akula_words_2020} test datasets, using the most popular and widely-used metrics, \ie accuracy~\cite{li_grounded_2022,liu_grounding_2023,dai_simvg_2024,xiao_florence-2_2024} and intersection over union (IoU)~\cite{suris_vipergpt_2023, ke_hydra_2024} . For the segmentation, we follow the evaluation protocol in previous works~\cite{rasheed_glamm_2024, zhang_psalm_2024} on the RefCOCOg and RefCLEF datasets~\cite{kazemzadeh_referitgame_2014} with IoU as metric. The IoU in segmentation\footnote{In some previous works~\cite{lai_lisa_2024, zhang_psalm_2024, rasheed_glamm_2024}, it is termed as ``cIoU".} is defined as the number of intersection pixels divide by the union pixels accumulated in the test set. In addition, we compare the efficiency of \methodname{} with other compositional baselines~\cite{ke_hydra_2024, suris_vipergpt_2023} in terms of runtime, LLM token usage, and financial cost.

\subsection{Quantitative Analysis}
\label{sec:quantitative}

\begin{table}[t]
\centering
\scalebox{0.83}{
\begin{tabular}{c|l|cccc}
\toprule[0.4mm]
\rowcolor{mygray} \textbf{} & \textbf{Method} & \textbf{Ref} & \textbf{Ref+} & \textbf{Refg} & \textbf{Ref-Adv} \\ \hline \hline
\multirow{11}{*}{\rotatebox[origin=c]{90}{End-to-End}} & GLIP-L~\cite{li_grounded_2022} & 55.0 & 51.1 & 54.6 & 55.7 \\
& KOSMOS-2~\cite{peng_kosmos-2_2023} & 57.4 & 50.7 & 61.7 & - \\
& YOLO-World-X~\cite{cheng_yolo-world_2024} & 12.1 & 12.1 & 32.9 & 32.2 \\
& YOLO-World-V2-X~\cite{cheng_yolo-world_2024} & 19.8 & 16.8 & 36.5 & 33.1 \\
& GroundingDINO-T~\cite{liu_grounding_2023} & 61.6 & 59.7 & 60.6 & 60.5 \\
& GroundingDINO-B~\cite{liu_grounding_2023} & 90.8 & 84.6 & 80.3 & 78.0 \\
& SimVG~\cite{dai_simvg_2024} & 94.9 & 91.0 & 88.9 & 74.4 \\
& Florence2-B~\cite{xiao_florence-2_2024} & 94.5 & 91.2 & 88.3 & 72.2 \\
& Florence2-L~\cite{xiao_florence-2_2024} & 95.1 & 92.5 & 90.9 & 71.8 \\
& GPT-4o-2024-05-13~\cite{openai_gpt-4o_2024} & 30.5 & 26.2 & - & - \\
& Qwen2.5-VL-72B~\cite{bai_qwen25-vl_2025} & 94.6 & 92.2 & 90.3 & - \\ \hline
\multirow{7}{*}{\rotatebox[origin=c]{90}{Compositional}} & Code-bison~\cite{stanic_towards_2024} & 44.4 & 38.2 & - & -\\
& ViperGPT†~\cite{suris_vipergpt_2023} & 62.6 & 62.3 & 67.2 & 60.7 \\
& HYDRA†~\cite{ke_hydra_2024} & 60.9 & 56.5 & 62.9 & 54.4 \\
& \methodname{}† & \textbf{70.1} & \textbf{64.1} & \textbf{69.5} & \textbf{65.1} \\ \cline{2-6}
& ViperGPT* & 68.6 & 73.8 & 68.7 & 58.2 \\
& HYDRA* & 65.7 & 66.2 & 59.9 & 48.3 \\
& \methodname{}* & \textbf{96.2} & \textbf{92.8} & \textbf{91.6} & \textbf{75.4} \\
\bottomrule[0.4mm]
\end{tabular}}
\vspace{-2.5mm}
\caption{\textbf{Quantitative comparison (accuracy) on referring expression detection task on RefCOCO, RefCOCO+, RefCOCOg~\cite{kazemzadeh_referitgame_2014} and Ref-Adv~\cite{akula_words_2020} set.} The training set of the GroundingDINO-B, SimVG and Florence2 includes RefCOCO/RefCOCO+/RefCOCOg datasets. 
Other E2E methods are zero-shot. \methodname{} is tested with two versions, one uses the same foundation models with HYDRA and ViperGPT for fair comparison, and another one uses the SoTA models for best performance. In compositional methods, they are grouped with the same VLMs for fair comparison. The methods† and methods* use the same VFMs (GLIP-L, BLIP) and same VFMs (Florence2-L, InternVL2), respectively. Both groups use GPT-4o Mini.}
\label{tab:detection}
\vspace{-4.5mm}
\end{table}

\noindent\textbf{Referring Expression Detection.} We conduct the experiments for quantitative comparison between \methodname{} and SoTA end-to-end (E2E) and compositional baselines on RefCOCO~\cite{kazemzadeh_referitgame_2014} test A, RefCOCO+~\cite{kazemzadeh_referitgame_2014} test A, RefCOCOg~\cite{kazemzadeh_referitgame_2014} test, and Ref-Adv~\cite{akula_words_2020} test set. As shown in \autoref{tab:detection}, \methodname{} demonstrates SoTA performance in comparison. To fairly compare to other compositional methods, \methodname{} use the same LLMs and foundation models (including GPT-4o Mini and GLIP~\cite{li_grounded_2022}) as HYDRA~\cite{ke_hydra_2024} and ViperGPT~\cite{ke_hydra_2024}, and performs better in all four datasets, highlighting the advantage of the design of \methodname{}. ViperGPT and HYDRA, even with SoTA models, fail to achieve significant improvement due to its reliance on LLMs for generating complex Python code, which becomes a bottleneck. Using the SoTA foundation models, \methodname{} performs better than the SoTA E2E methods, as well as compositional approaches such as HYDRA and ViperGPT, \czx{because \methodname{} is equipped with self-correction mechanism and logic reasoning, also breaks down the complex reasoning into simpler subtasks, which are easier to be solved by LLMs and VLMs.} 
For more comprehensive results and comparison, please see \autoref{tab:detection_supp} in supplementary material including more methods and metrics.

\begin{table}[t]
\centering
\scalebox{0.88}{
\begin{tabular}{c|l|cc}
\toprule[0.4mm]
\rowcolor{mygray} \textbf{} & \textbf{} & \multicolumn{2}{c}{\textbf{RefCOCOg}} \\ 
\rowcolor{mygray} \textbf{} & \textbf{Method} & \textbf{Val} & \textbf{Test} 
\\ \hline \hline
\multirow{7}{*}{\rotatebox[origin=c]{90}{E2E (not VLM)}} 
& GLIP-L + SAM~\cite{li_grounded_2022, kirillov_segment_2023} & 35.6 & 38.4 \\ 
& MCN~\cite{luo_multi-task_2020} & 49.2 & 49.4 \\
& VLT~\cite{ding_vision-language_2021} & 55.0 & 57.7 \\
& CRIS~\cite{wang_cris_2022} & 59.9 & 60.4 \\
& LAVT~\cite{yang_lavt_2022} & 61.2 & 62.1 \\
& GRES~\cite{liu_gres_2023} & 65.0 & 66.0 \\
& GroundingDINO-B + SAM~\cite{liu_grounding_2023, kirillov_segment_2023} & 58.7 & 61.4 \\ \hline
\multirow{5}{*}{\rotatebox[origin=c]{90}{E2E (VLM)}} & LISA-7B~\cite{lai_lisa_2024} & 66.4 & 68.5 \\
& PerceptionGPT~\cite{pi_perceptiongpt_2024} & 70.7 & 71.9 \\
& GSVA~\cite{xia_gsva_2024} & 73.2 & 73.9 \\
& PSALM~\cite{zhang_psalm_2024} & 73.8 & 74.4 \\ 
& GLaMM~\cite{rasheed_glamm_2024} & 74.2 & 74.9 \\ \hline
\multirow{6}{*}{\rotatebox[origin=c]{90}{Compositional}} & ViperGPT†~\cite{suris_vipergpt_2023} & 46.9 & 44.1 \\
& HYDRA†~\cite{ke_hydra_2024} & 51.9 & 53.1 \\
& \methodname{}† & \textbf{54.4} & \textbf{54.6}\\ \cline{2-4}
& ViperGPT* & 50.9 & 44.8 \\
& HYDRA* & 53.2 & 54.0 \\
& \methodname{}* & \textbf{76.4} & \textbf{76.0} \\
\bottomrule[0.4mm]
\end{tabular}}
\vspace{-2.5mm}
\caption{\textbf{Quantitative comparison (IoU) on Referring Expression Segmentation task on RefCOCOg val and test set~\cite{kazemzadeh_referitgame_2014}.} The training data of all E2E methods except GLIP-L include RefCOCOg dataset. We compare \methodname{} with two compositional baselines using different VFMs. Each method is evaluated in two versions: (†) uses the same VFMs as the original HYDRA and ViperGPT for a fair comparison; (*) uses the SoTA VFMs for best performance. All use GPT-4o Mini as the LLMs.
}
\label{tab:segmentation}
\vspace{-2.5mm}
\end{table}

\begin{table}[t]
\centering
\scalebox{0.88}{
\begin{tabular}{c|l|cc}
\toprule[0.4mm]
\rowcolor{mygray} \textbf{} &  & \multicolumn{2}{c}{\textbf{RefCLEF}} \\ 
\rowcolor{mygray} \textbf{} & \textbf{Method} & \textbf{Test A} & \textbf{Test B} \\ \hline \hline
\multirow{3}{*}{\rotatebox[origin=c]{90}{E2E}} 
& GLIP-L + SAM~\cite{li_grounded_2022, kirillov_segment_2023} & 50.9 & 56.4 \\
& GroundingDINO-B + SAM~\cite{liu_grounding_2023, kirillov_segment_2023} & 62.5 & 63.1 \\
& GLaMM~\cite{rasheed_glamm_2024} & 67.5 & 66.8 \\ \hline
\multirow{6}{*}{\rotatebox[origin=c]{90}{Compositional}} & ViperGPT†~\cite{suris_vipergpt_2023} & 55.6 & 56.9 \\
& HYDRA†~\cite{ke_hydra_2024} & 56.4 & 58.1 \\
& \methodname{}† & \textbf{58.1} & \textbf{59.3} \\ \cline{2-4}
& ViperGPT* & 61.2 & 58.3 \\
& HYDRA* & 56.5 & 58.3 \\
& \methodname{}* & \textbf{68.8} & \textbf{67.4} \\
\bottomrule[0.4mm]
\end{tabular}}
\vspace{-2.5mm}
\caption{\textbf{Quantitative comparison (IoU) on Referring Expression Segmentation task on RefCLEF~\cite{kazemzadeh_referitgame_2014}.} The training data of all E2E methods except GLIP-L include RefCOCOg dataset. The configurations (†, *) of compositional methods follow \autoref{tab:segmentation}.}
\label{tab:segmentation2}
\vspace{-4.5mm}
\end{table}

\noindent\textbf{Referring Expression Segmentation.} Results on RefCOCOg and RefCLEF~\cite{kazemzadeh_referitgame_2014} are shown in \autoref{tab:segmentation} and \autoref{tab:segmentation2}. In both datasets, \methodname{} demonstrates higher performance compared to SoTA E2E baselines and other compositional baselines~\cite{ke_hydra_2024, suris_vipergpt_2023} using the same foundation models and language models. Consistent with the observation from \autoref{tab:detection}, replacing the VFMs in compositional methods with the best performing models, \ie GLaMM and InternVL2, incrementally improve their performances in segmentation. These experiments show that \methodname{} is capable of utilizing best performing VLM-based methods and yet achieve a better performance in this task.

\noindent\textbf{Performance Analysis.} The previous compositional baselines~\cite{suris_vipergpt_2023, ke_hydra_2024} compute the grounding performance excluding the runtime errors caused by LLM's uncontrollable output. The scores reported in the benchmarks are not reliable in real applications due to the high failure rate. We conduct a fair comparison with previous compositional baselines~\cite{ke_hydra_2024, suris_vipergpt_2023} by using the same LLM (GPT-4o Mini~\cite{openai_gpt-4o_2024}), grounding model (GLIP-Large~\cite{li_grounded_2022}, and SAM~\cite{kirillov_segment_2023}) in the segmentation task. We report the IoU scores including/excluding the runtime errors and the runtime failure rate in  \autoref{tab:failure_rate}. The original ViperGPT lacks the validation in inference, leading to a high failure rate. 
HYDRA reduces the error rate by introducing incremental reasoning and a controller to validate the plan, but still reports a 10.3\% failure rate. Our proposed method \methodname{} produces a more reliable pipeline (0.3\% runtime failure rate) with the validation for the full-chain inference, where each state verifies the intermediate outputs. We find the main failure reason is due to the ChatGPT content policy which misclassifies some queries such as ``man cut off'' as violent text and fails to %
respond, however this query asks for a man only half of whose body appears in the image. Furthermore, comparing the performance ignoring the runtime error, we still observe a noticeable improvement of \methodname{} compared to the baselines with the same foundation models.


\begin{table}[t]
\centering
\scalebox{0.9}{
\begin{tabular}{l|ccc}
\toprule[0.4mm]
\rowcolor{mygray} \textbf{} & \textbf{Failure} & \textbf{RefCOCOg} & \textbf{RefCOCOg} \\
\rowcolor{mygray} \textbf{Method} & \textbf{(\%)} & \textbf{(Exc. Err.)} & \textbf{(Inc. Err.)} \\\hline \hline
ViperGPT~\cite{suris_vipergpt_2023} & 70.2 & 44.1 & 15.3 \\
HYDRA~\cite{ke_hydra_2024} & 10.3 & 53.1 & 47.6 \\
\methodname{} (Ours) & \textbf{00.3} & \textbf{54.6} & \textbf{54.4} \\
\bottomrule[0.4mm]
\end{tabular}}
\vspace{-2.5mm}
\caption{\textbf{Performance comparison with compositional methods.} All results are evaluated on the RefCOCOg (IoU) test set for segmentation. All methods use the same foundation models.}
\label{tab:failure_rate}
\vspace{-2.5mm}
\end{table}

\subsection{Ablation Studies}
\label{sec:ablation}

\noindent\textbf{Contribution of Modules.} To examine the contributions of key components, \emph{deterministic finite-state automaton} (DFA), logic \czx{and answerer}, we conduct ablation studies on RefCOCO, RefCOCO+, RefCOCOg and Ref-Adv dataset shown in \autoref{tab:ablation}. To compare in detail the contribution of each component, we examined different combinations of the system:  \textit{(1) Disabling DFA:} Disabling the self-correction transitions (\textcolor{red}{red arrows} in \autoref{fig:pipeline}); \textit{(2) Disabling Logic:} The entities detected in the \emph{Perception} state are redirected to the \emph{Answerer} bypassing the \emph{Logic Generation} and \emph{Logic Reasoning}; \textit{(3) Disabling Answerer:} the answer is not validated by the answerer; and the feasible combinations of (1), (2) and (3). 
\czx{Comparing rows 1 and 4, 3 and 6, 5 and 7, the DFA self-correction mechanism provides substantial performance improvement.
Comparing rows 1 and 3, 2 and 5, the enabling of logic leads to a significant improvement in the performance. 
Comparing rows 1 and 2, 3 and 5, 6 and 7, the answerer validates the result and improves the accuracy.}
From row 7, enabling all DFA self-correction, logic and answerer improves the performance and reaches the SoTA. 


\noindent\textbf{Selection of Grounding Models.} We conduct experiments using different grounding models as the entity detector in \methodname{}. We tested \methodname{} with various grounding methods~\cite{li_grounded_2022, liu_grounding_2023, dai_simvg_2024, rasheed_glamm_2024}
on RefCOCO for detection and RefCOCOg for segmentation, as shown in \autoref{tab:grounding_comparison}. As \methodname{} is flexible in choosing foundation models, \methodname{} can gain performance by updating to SoTA grounding models. Comparing the performance of original grounding method with \methodname{} employing it, the performance is usually improved when using \methodname{}.



\begin{table}[t]
\centering
\scalebox{0.78}{
\begin{tabular}{ccc|cccc}
\toprule[0.4mm]
\rowcolor{mygray} \textbf{DFA} & \textbf{Logic} & \textbf{Answerer} & \textbf{Ref} & \textbf{Ref+} & \textbf{Refg} & \textbf{Ref-Adv} \\\hline \hline
\ding{55} & \ding{55} & \ding{55} & 51.7 & 51.3 & 48.8 & 34.1 \\
\ding{55} & \ding{55} & \ding{51} & 57.4 & 57.6 & 50.8 & 34.3 \\
\ding{55} & \ding{51} & \ding{55} & 57.8 & 62.0 & 51.3 & 36.4 \\
\ding{51} & \ding{55} & \ding{55} & 68.5 & 69.3 & 70.1 & 64.7 \\
\ding{55} & \ding{51} & \ding{51} & 78.1 & 68.4 & 53.4 & 43.6 \\
\ding{51} & \ding{51} & \ding{55} & 76.2 & 73.8 & 75.8 & 69.7 \\
\ding{51} & \ding{51} & \ding{51} & \textbf{96.2} & \textbf{92.8} & \textbf{91.6} & \textbf{75.4}\\
\bottomrule[0.4mm]
\end{tabular}}
\vspace{-2.5mm}
\caption{\textbf{Contribution of components of \methodname{}.} Results (accuracy) are based on the RefCOCO test A, RefCOCO+ test A, RefCOCOg test and Ref-Adv test set.}
\label{tab:ablation}
\vspace{-2.5mm}
\end{table}

\noindent\textbf{Selection of LLM.} We conduct experiments with different LLMs, including LLaMA3.1~\cite{dubeyLlama2024}, Gemma2~\cite{team_gemma_2024}, DeepSeek-R1~\cite{deepseek-ai_deepseek-r1_2025-1}, and OpenAI LLMs~\cite{openai_gpt-4o_2024}, evaluating them on the RefCOCO and RefCOCOg dataset in \autoref{tab:llm_comparison}. The results indicate that for the easier RefCOCO dataset, the performance differences between LLMs are minimal. However, for the more challenging RefCOCOg dataset, the performance gap becomes more significant. Given these findings, a smaller high-efficiency LLM, such as GPT-4o Mini, is preferred, as it maintains competitive performance while achieving significantly lower inference time.


\begin{table}[t]
\centering
\scalebox{0.8}{
\begin{tabular}{l|cc}
\toprule[0.4mm]
\rowcolor{mygray} & \multicolumn{2}{c}{\textbf{RefCOCO}} \\ 
\rowcolor{mygray} \textbf{Method} & Itself & In \methodname{} \\ \hline \hline
GLIP-L~\cite{li_grounded_2022} & 55.0 & \textbf{70.1} \\
GroundingDINO-T~\cite{liu_grounding_2023} & 61.6 & \textbf{64.9} \\
GroundingDINO-B~\cite{liu_grounding_2023} & 90.8 & \textbf{91.7} \\
SimVG~\cite{dai_simvg_2024} & 94.9 & \textbf{95.4} \\
Florence2-L~\cite{xiao_florence-2_2024} & 95.1 & \textbf{96.2}\\\hline
\rowcolor{mygray} \textbf{Method} & \multicolumn{2}{c}{\textbf{RefCOCOg}} \\\hline
GLIP+SAM~\cite{li_grounded_2022, kirillov_segment_2023} & 38.4 & \textbf{54.6} \\
GroundingDINO-B+SAM~\cite{liu_grounding_2023, kirillov_segment_2023} & 61.4 & \textbf{65.0} \\
GLaMM~\cite{rasheed_glamm_2024} & 74.9 & \textbf{76.0} \\
\bottomrule[0.4mm]
\end{tabular}}
\vspace{-2.5mm}
\caption{\textbf{Comparison of the performance using different grounding methods for \methodname{}.} All results are evaluated on the RefCOCO (accuracy) for detection and RefCOCOg (IoU) for segmentation. The column \emph{Itself} shows the performance of the methods, and column \emph{In \methodname{}} shows the performance of NAVER using these methods for grounding.}
\label{tab:grounding_comparison}
\vspace{-2.5mm}
\end{table}

\begin{table}[t]
\centering
\scalebox{0.8}{
\begin{tabular}{lc|cc|cc}
\toprule[0.4mm]
\rowcolor{mygray} & & \multicolumn{2}{c|}{\textbf{RefCOCO}} & \multicolumn{2}{c}{\textbf{RefCOCOg}} \\
\rowcolor{mygray} \textbf{LLM} & \textbf{Size} & \textbf{Time (Sec.)} & \textbf{Acc.} & \textbf{Time (Sec.)} & \textbf{Acc.} \\\hline \hline
DeepSeek-R1~\cite{deepseek-ai_deepseek-r1_2025-1} & 70B & 71.27 & 95.4 & 82.13 & 85.8\\
LLaMA3.1~\cite{dubeyLlama2024} & 70B & 12.41 & 96.0 & 19.40 & 88.8 \\
LLaMA3.1~\cite{dubeyLlama2024} & 8B & 8.23 & 95.0 & 9.25 & 90.5 \\
Gemma2~\cite{team_gemma_2024} & 3B & 8.18 & 96.1 & 9.13 & 91.1 \\
GPT-4o~\cite{openai_gpt-4o_2024} & - & 8.67 & \textbf{96.6} & 10.29 & 89.7 \\
GPT-4o Mini~\cite{openai_gpt-4o_2024} & - & 5.74 & 96.2 & 7.91 & \textbf{91.6} \\
\bottomrule[0.4mm]
\end{tabular}}
\vspace{-2.5mm}
\caption{\textbf{Comparison of LLM selection for \methodname{}.} Results are evaluated on the RefCOCO and RefCOCOg for detection. The time values show the average seconds spent on each data sample.}
\label{tab:llm_comparison}
\vspace{-2.5mm}
\end{table}
\section{Conclusion}
This paper introduces a Neuro-Symbolic Compositional Automaton for Visual Grounding with Explicit Logic Reasoning (\methodname{}), a novel method that overcomes limitations in previous works by combining explicit logic reasoning, and self-correcting mechanisms as a finite-state automaton. This design enhances \methodname{}’s capacity to address complex visual queries with improved accuracy, robustness, and interpretability. Extensive experiments on public benchmarks confirm that \methodname{} achieves state-of-the-art performance, providing a more interpretable and robust solution for reasoning-based visual grounding tasks.

\noindent\textbf{Broader Impact.} The overall design of \methodname{}, using an automaton to combine perception and logic generation and reasoning to provide robust and interpretable answers seems applicable to many other complex query answering problems. \methodname{} makes use of the fact that LLMs are better at confirming that a given answer satisfies a complex query, than finding the answer itself; and uses this, together with other methods, to self-correct. 

\noindent\textbf{Limitation.} The logic code generation is limited by the capabilities of the LLMs. Enhancing LLMs for code generation can be the future direction.

\section*{Acknowledgments}
This research is sponsored by the DARPA Assured Neuro Symbolic Learning and Reasoning (ANSR) program under award number FA8750-23-2-1016.

{
    \small
    \bibliographystyle{ieeenat_fullname}
    \bibliography{references}
}

\clearpage
\setcounter{page}{1}
\maketitlesupplementary

\noindent This supplementary material provides additional details about the proposed method \methodname{}. In particular, it provides details about 
the additional quantitative comparison result with both accuracy and IoU in \autoref{sec:supp_quantitative},
the complexity analysis in \autoref{sec:complexity_analysis},
the ablation studies for VLM in \autoref{sec:supp_vlm}, 
the analysis of self-correction mechanisms in \autoref{sec:supp_self_correction}, 
the performance analysis for query length in \autoref{sec:supp_query_length},
the ablation studies for captioner in \autoref{sec:ablation_captioner},
and the prompts used for the LLMs and VLMs in \autoref{sec:supp_prompt}.

\begin{table*}[t]
\centering
\scalebox{0.83}{
\begin{tabular}{c|l|cc|cc|cc|cc}
\toprule[0.4mm]
\rowcolor{mygray} \textbf{} & & \multicolumn{2}{c|}{\textbf{RefCOCO}} & \multicolumn{2}{c|}{\textbf{RefCOCO+}} & \multicolumn{2}{c|}{\textbf{RefCOCOg}} & \multicolumn{2}{c}{\textbf{Ref-Adv}} \\ 
\rowcolor{mygray} \textbf{} & \textbf{Method} & \textbf{Acc.} & \textbf{IoU} & \textbf{Acc.} & \textbf{IoU} & \textbf{Acc.} & \textbf{IoU} & \textbf{Acc.} & \textbf{IoU} \\\hline \hline
\multirow{9}{*}{\rotatebox[origin=c]{90}{E2E}} & GLIP-L~\cite{li_grounded_2022} & 55.0 & 54.1 & 51.1 & 51.3 & 54.6 & 54.8 & 55.7 & 55.2 \\
& KOSMOS-2~\cite{peng_kosmos-2_2023} & 57.4 & - & 50.7 & - & 61.7 & - & - & - \\ 
& YOLO-World-X~\cite{cheng_yolo-world_2024} & 12.1 & 12.7 & 12.1 & 12.7 & 32.9 & 33.8 & 32.2 & 34.2 \\
& YOLO-World-V2-X~\cite{cheng_yolo-world_2024} & 19.8 & 20.0 & 16.8 & 17.2 & 36.5 & 37.3 & 33.1 & 34.8 \\
& GroundingDINO-T~\cite{liu_grounding_2023} & 61.6 & 60.2 & 59.7 & 58.9 & 60.6 & 59.7 & 60.5 & 59.8 \\
& GroundingDINO-B~\cite{liu_grounding_2023} & 90.8 & 85.2 & 84.6 & 77.5 & 80.3 & 69.4 & 78.0 & 73.1 \\
& SimVG~\cite{dai_simvg_2024} & 94.9 & 86.9 & 91.0 & 83.9 & 88.9 & 81.3 & 74.4 & 70.7 \\
& Florence2-B~\cite{xiao_florence-2_2024} & 94.5 & 89.5 & 91.2 & 86.5 & 88.3 & 85.0 & 72.2 & 71.9 \\
& Florence2-L~\cite{xiao_florence-2_2024} & 95.1 & 90.6 & 92.5 & 88.2 & 90.9 & 87.6 & 71.8 & 71.8 \\ \hline
\multirow{10}{*}{\rotatebox[origin=c]{90}{Compositional}} & Code-bison~\cite{stanic_towards_2024} & 44.4 & - & 38.2 & - & - & - & - & - \\
& ViperGPT†~\cite{suris_vipergpt_2023} & 62.6 & 59.4 & 62.3 & 58.7 & 67.2 & 63.6 & 60.7 & 58.6 \\
& HYDRA†~\cite{ke_hydra_2024} & 60.9 & 58.2 & 56.5 & 54.9 & 62.9 & 60.8 & 54.4 & 53.5 \\
& \methodname{}† & \textbf{70.1} & \textbf{67.9} & \textbf{64.1} & \textbf{63.8} & \textbf{69.5} &  \textbf{59.2} & \textbf{65.1} & \textbf{63.4} \\ \cline{2-10}
& ViperGPT‡ & 67.1 & 64.0 & 73.2 & 68.8 & 65.6 & 63.5 & 60.1 & 59.5 \\
& HYDRA‡ & 63.5 & 61.3 & 62.9 & 60.9 & 59.8 & 58.4 & 53.2 & 53.5 \\
& \methodname{}‡ & \textbf{91.7} & \textbf{83.3} & \textbf{82.4} & \textbf{78.4} & \textbf{75.9} & \textbf{72.5} & \textbf{87.3} & \textbf{74.2} \\ \cline{2-10}
& ViperGPT* & 68.6 & 66.5 & 73.8 & 70.4 & 68.7 & 66.8 & 58.2 & 58.3 \\
& HYDRA* & 65.7 & 64.6 & 66.2 & 65.3 & 59.9 & 60.5 & 48.3 & 51.8 \\
& \methodname{}* & \textbf{96.2} & \textbf{91.7} & \textbf{92.8} & \textbf{88.4} & \textbf{91.6} & \textbf{88.1} & \textbf{75.4} & \textbf{75.3} \\
\bottomrule[0.4mm]
\end{tabular}}
\caption{\textbf{Accuracy and IoU performance on the referring expression detection task.} Results are shown on the RefCOCO, RefCOCO+, RefCOCOg~\cite{kazemzadeh_referitgame_2014}, and Ref-Adv~\cite{akula_words_2020} datasets. In compositional methods, they are grouped with the same VLMs for fair comparison. The methods with same symbols (†, ‡, *) use the same VFMs. The VFMs used are: (†) uses GLIP-L and BLIP; (‡) uses GroundingDINO-B and InternVL2; (*) uses Florence2-L and InternVL2, respectively. All groups use GPT-4o Mini.}
\label{tab:detection_supp}
\end{table*}

\section{Additional Quantitative Comparison}
\label{sec:supp_quantitative}

In this section, we report both the accuracy and Intersection over Union (IoU) performance for the referring expression detection task. \autoref{tab:detection_supp} presents a quantitative comparison on the RefCOCO, RefCOCO+, RefCOCOg~\cite{kazemzadeh_referitgame_2014}, and Ref-Adv~\cite{akula_words_2020} datasets. Additionally, we include results for the compositional methods using GroundingDINO as the VFM, marked with ‡. We observed that for each configuration, our proposed method \methodname{} outperforms the other compositional methods using the same foundation models and also outperform the grounding methods themselves. These results further validate that our approach consistently achieves superior IoU performance over existing baselines.

\section{Complexity Analysis}
\label{sec:complexity_analysis}
Efficiency results for \methodname{} and compositional baselines are provided in \autoref{tab:complexiy}, which includes average runtime, token usage, and cost per sample. For a fair comparison, we use the same foundation models and LLM for the experiments. From the results, we observe \methodname{} is more efficient than HYDRA~\cite{ke_hydra_2024} with higher performance and lower runtime failure rate. Although ViperGPT~\cite{suris_vipergpt_2023} has lower time complexity, this advantage is offset by reduced performance due to its lack of validation. These results demonstrate that \methodname{}'s flexible automaton design reduces unnecessary computations by dynamically adapting the pipeline.

\begin{table}[h]
\centering
\scalebox{0.9}{
\begin{tabular}{l|cccc}
\toprule[0.4mm]
\rowcolor{mygray} \textbf{} & \textbf{Time} & \multicolumn{2}{c}{\textbf{\# LLM Tokens}} & \textbf{LLM Cost} \\
\rowcolor{mygray} \textbf{Method} & \textbf{(Seconds)} & \textbf{Input} & \textbf{Output} & \textbf{(USD)} \\\hline \hline
ViperGPT~\cite{suris_vipergpt_2023} & \textbf{2.02} & 4169 & \textbf{29} & 0.00064 \\
HYDRA~\cite{ke_hydra_2024} & 14.93 & 19701 & 618 & 0.00332  \\
\methodname{} (Ours) & 5.96 & \textbf{2083} & 39 & \textbf{0.00034} \\
\bottomrule[0.4mm]
\end{tabular}}
\caption{\textbf{Complexity comparison between compositional methods.} All results are evaluated on the RefCOCOg test set with the GPT-4o Mini for fair comparison. All values are the average per data sample. All methods use the same foundation models.}
\label{tab:complexiy}
\end{table}

\section{Ablation Studies for VLMs}
\label{sec:supp_vlm}

We evaluate the impact of four different Vision-Language Models (VLMs)~\cite{li_blip-2_2023, xue_xgen-mm_2024, liu_improved_2023, chen_internvl_2024} on the performance of \methodname{}. The results are shown in \autoref{tab:ablation_vlm}. All tested models achieve comparable results on both RefCOCO and RefCOCO+. 
Among the four VLMs, InternVL2-8B achieves slightly better performance with accuracy scores of 96.2 on RefCOCO and 92.8 on RefCOCO+. However, the differences between the VLMs are marginal, indicating that the architecture of \methodname{} is insensitive to the choice of VLM. This shows \methodname{}'s ability to decompose complex tasks into simpler subtasks, reducing the reliance on the capabilities of the VLMs.
Given the similar performance across all tested VLMs, we select InternVL2-8B for its availability and accessibility, ensuring simpler implementation without compromising performance. This demonstrates the flexibility of \methodname{} in adopting different VLMs while delivering state-of-the-art performance.


\begin{figure*}[t]
\centering
\includegraphics[width=0.7\linewidth]{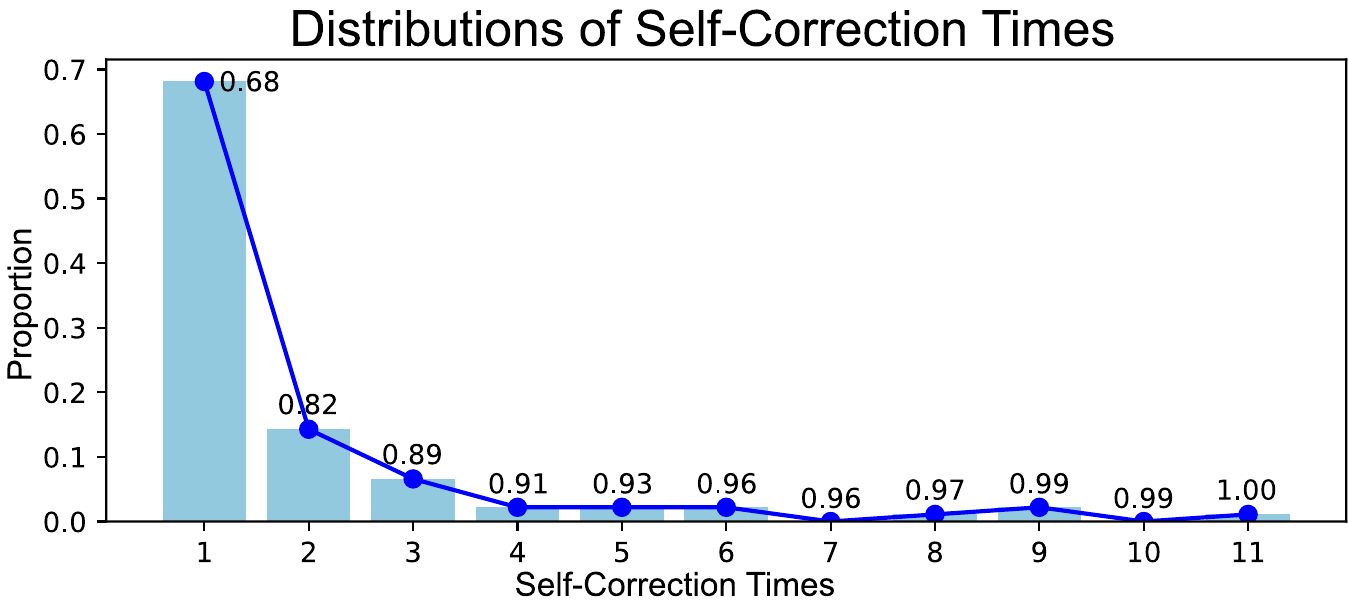}
\caption{\textbf{Distribution of self-correction times in \methodname{}.} The x-axis represents the number of retries for self-correction, while the y-axis shows the proportion of samples requiring self-correction within each group of self-correction times. The annotated values indicate the cumulative proportion of samples resolved by that number of retries.}
\label{fig:error}
\end{figure*}

\begin{figure*}[t]
\centering
\includegraphics[width=0.7\linewidth]{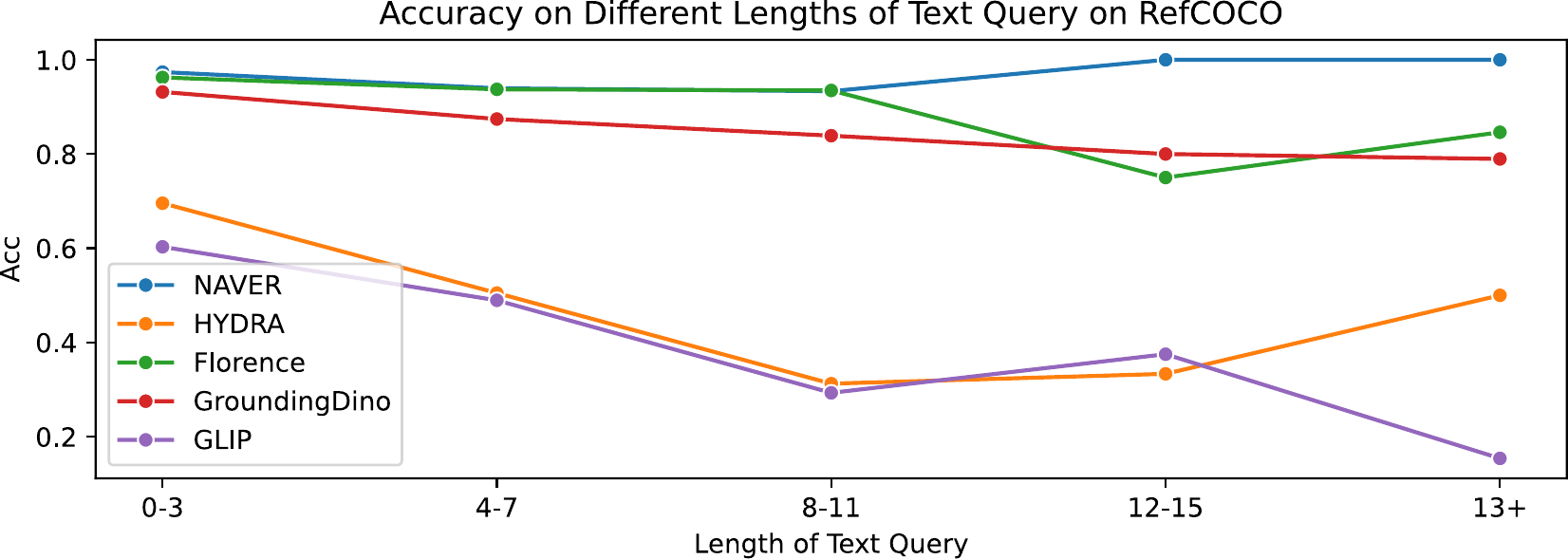}
\caption{\textbf{Accuracy performance of different text query length in \methodname{}.} The x-axis represents the length of text query, while the y-axis shows the accuracy of \methodname{} and baselines within each group of query length.}
\label{fig:query_length}
\end{figure*}

\begin{table}[t]
\centering
\scalebox{1}{
\begin{tabular}{l|cc}
\toprule[0.4mm]
\rowcolor{mygray} \textbf{VLM} & \textbf{RefCOCO} & \textbf{RefCOCO+} \\\hline \hline
BLIP2-XXL~\cite{li_blip-2_2023} & 95.4 & 91.7 \\
xGen-MM~\cite{xue_xgen-mm_2024} & 94.5 & 87.9 \\
LLaVA1.5-7B~\cite{liu_improved_2023} & 95.5 & 91.9 \\
InternVL2-8B~\cite{chen_internvl_2024} & \textbf{96.2} & \textbf{92.8} \\
\bottomrule[0.4mm]
\end{tabular}}
\caption{\textbf{Comparison of VLM selection for \methodname{}.} All results (accuracy) are evaluated on the RefCOCO and RefCOCO+ testA datasets~\cite{kazemzadeh_referitgame_2014}.}
\label{tab:ablation_vlm}
\end{table}

\begin{table*}[t]
\centering
\scalebox{1}{
\begin{tabular}{cc|cccc}
\toprule[0.4mm]
\rowcolor{mygray} \textbf{Captioner} & \textbf{ECE} & \textbf{RefCOCO} & \textbf{RefCOCO+} & \textbf{RefCOCOg} & \textbf{RefAdv} \\ \hline \hline
\ding{55} & VLM & 91.6 & 87.5 & 80.2 & 65.6 \\ 
VLM & LLM & \textbf{96.2} & \textbf{92.8} & \textbf{91.6} & \textbf{75.4}  \\
\bottomrule[0.4mm]
\end{tabular}
}
\caption{\textbf{Ablations for the captioner and type of entity category extractor (ECE).} All results are accuracy for referring expression detection task.}
\label{tab:ablation_captioner}
\end{table*}

\section{Self-Correction Analysis}
\label{sec:supp_self_correction}

We analyze the performance of the self-correction mechanism in \methodname{}, focusing on its ability to address errors during inference. In this mechanism, each time the system transitions into the self-correction state (indicated by a \textcolor{red}{red arrow} in \autoref{fig:pipeline}), it is counted as one retry. Our experiments, conducted on the RefCOCO test A dataset, show that approximately 10\% of the samples require self-correction.
To better understand the behavior of the mechanism, we calculate the proportion of samples resolved after each number of retries within the subset of data requiring self-correction. These results are visualized as a bar chart in \autoref{fig:error}. The analysis shows that a single retry resolves 68\% of the errors, and 96\% of the errors are resolved within six retries.
To prevent infinite looping, we limit the maximum number of retries to six, as this threshold effectively addresses the majority of the errors.

\section{Query Length Analysis}
\label{sec:supp_query_length}

We analyze the impact of text query length on performance, comparing \methodname{} with baselines across different query lengths. The results, shown in \autoref{fig:query_length}, indicate that \methodname{} consistently reaches SoTA performance compared to all baselines regardless of query length. While most baseline methods experience a performance decline as query length increases, the extent of decrease varies. Notably, older models like GLIP~\cite{li_grounded_2022} show a significant performance drop for longer queries, suggesting difficulties in handling complex text inputs. In contrast, \methodname{} maintains stable performance, showing its robustness in processing longer and more complex queries.

\section{Ablations for Captioner}
\label{sec:ablation_captioner}

In the perception state, \methodname{} first converts the image into a rich caption and then lets an LLM‑based entity‑category extractor (ECE) decide which object classes are relevant to the query.  An intuitive alternative is to skip captioning and ask the VLM to predict categories directly. \autoref{tab:ablation_captioner} shows that this seemingly simpler choice causes a consistent accuracy drop of 4.6\% on RefCOCO, 5.3\% on RefCOCO+, 11.4\% on RefCOCOg, and 9.8\% on Ref‑Adv.  The larger gap on the more complex datasets (RefCOCOg, Ref‑Adv) suggests that a textual scene summary helps the downstream logic generator reason about complex descriptions. The clear improvement brought by the Captioner justifies keeping it in the final system.


\section{LLM and VLM Prompts}
\label{sec:supp_prompt}

\methodname{} employs LLMs and VLMs in five different roles as described in \autoref{sec:method}: as a caption generator VLM in the \emph{Perception} state, an entity extractor LLM in the \emph{Perception} state, a logic query generator LLM in the \emph{Logic Generation} state, a relation recognizer VLM in the \emph{Logic Generation} state, and an answerer in the \emph{Answering} state. The prompts utilized for each role are carefully crafted to generate precise outputs aligned with the requirements of each state, ensuring accurate visual grounding and robust reasoning. Prompts~\hyperref[prompt:caption]{12.1} to \hyperref[prompt:answerer]{12.5} show the detailed prompt templates for each role.

\begin{itemize}
    \item Prompt~\hyperref[prompt:caption]{12.1}: This prompt instructs the VLM to generate descriptive captions $I_c$ for the input image $I$. The generated captions represent the visual content in textual form, which serves as input for the entity extractor LLM in subsequent processing.

    \item Prompt~\hyperref[prompt:entity_extractor]{12.2}: This prompt is utilized in the \emph{Perception} state to guide the LLM in identifying entity categories $C$ relevant to the query $Q$ and the caption $I_c$. The output is a list of entity categories, which forms the basis for visual reasoning in the subsequent states.

    \item Prompt~\hyperref[prompt:logic_query]{12.3}: This prompt guides the logic query generator LLM to convert the textual query $Q$ into a ProbLog logic query in the \emph{Logic Generation} state. The output logic query is then utilized in the ProbLog interpreter to perform probabilistic logic reasoning.

    \item Prompt~\hyperref[prompt:relation]{12.4}: This prompt guides the relation recognizer VLM to identify relations $R$ between entities $E$ in the \emph{Logic Generation} state. These recognized relations are used to construct logic expressions, as the foundation for probabilistic logic reasoning.

    \item Prompt~\hyperref[prompt:answerer]{12.5}: This prompt is used in the \emph{Answering} state to guide the answerer VLM in validating whether the identified target $Y_{L1}$ satisfies the conditions of the query $Q$. It requests a binary response (``Yes'' or ``No'') to confirm whether the top-ranked candidate from the \emph{Logic Reasoning} state fulfills all query requirements, ensuring only valid results are returned as final outputs.
\end{itemize}

\begin{prompt}[title={Prompt \thetcbcounter: Captioner VLM}] \label{prompt:caption}
\textlangle image\textrangle Please describe the image in detail.
\end{prompt}

\begin{prompt}[title={Prompt \thetcbcounter: Entity Extractor LLM}] \label{prompt:entity_extractor}
You're an AI assistant designed to find detailed information from image.
\\ \\
You need to find important objects based on the given query which is the object you need to find. The query normally is a set of words which includes a object name and the attributes of the object.
\\ \\
Here are some examples:\\
Query: \textlangle example query 1\textrangle \\
Answer: \textlangle example answer 1\textrangle \\

Query: \textlangle example query 2\textrangle \\
Answer: \textlangle example answer 2\textrangle \\
...

Your output must be a JSON object contains the flatten list of string. For example: {"output": ["apple", "orange", "chair", "umbrella"]}\\

Caption: \textlangle caption\textrangle \\
Query: \textlangle query\textrangle \\
Answer: 
\end{prompt}

\begin{prompt}[title={Prompt \thetcbcounter: Logic Query Generator LLM}] \label{prompt:logic_query}
You're an AI assistant designed to generate the ProbLog code (a logic programming language similar to Prolog). \\

You need to generate a new rule "target" that will be used to query the target objects in the image based on given text prompt. \\

The names of entity categories are \textlangle entity categories\textrangle. \\

The output is the code. For example: \\
\textasciigrave\textasciigrave\textasciigrave problog \\
target(ID) :- entity(ID, "<some category>", \_, \_, \_, \_), relation(ID, \_, \_), attribute(ID, \_). \\
\textasciigrave\textasciigrave\textasciigrave \\

More examples: \\
find the target "\textlangle example query 1\textrangle"\\
\textasciigrave\textasciigrave\textasciigrave problog\\
\textlangle example ProbLog code 1\textrangle\\
\textasciigrave\textasciigrave\textasciigrave\\
\\
find the target "\textlangle example query 2\textrangle"\\
\textasciigrave\textasciigrave\textasciigrave problog\\
\textlangle example ProbLog code 2\textrangle\\
\textasciigrave\textasciigrave\textasciigrave\\
...\\

Complete the following ProbLog code:\\
\textasciigrave\textasciigrave\textasciigrave problog\\
\textlangle ProbLog code of context\textrangle\\
\textasciigrave\textasciigrave\textasciigrave \\

Your output should be the ProbLog code.

find the target "\textlangle query\textrangle"\\
Your answer: 
\end{prompt}

\begin{prompt}[title={Prompt \thetcbcounter: Relation Recognizer VLM}] \label{prompt:relation}
\textlangle image\textrangle You're an AI assistant designed to find the relations of objects in the given image.\\

The interested objects are highlighted by bounding boxes (X1, Y1, X2, Y2). They are:\\

A: the \textlangle category of A\textrangle labeled by red bounding box \textlangle bbox of A\textrangle.\\
B: the \textlangle category of B\textrangle labeled by red bounding box \textlangle bbox of B\textrangle.\\

Only consider the camera view. Note you are focusing to analyze the relation A to B, do not consider the relation B to A. Please answer "Yes" or "No" for the following question.\\

Is A \textlangle relation\textrangle{} B?\\

Your answer is:
\end{prompt}

\begin{prompt}[title={Prompt \thetcbcounter: Answerer VLM}] \label{prompt:answerer}
\textlangle image\textrangle You're an image analyst designed to check if the highlighted object in the image meets the query description.\\

The query is: "\textlangle query\textrangle"\\

Please check the highlighted object "A" in the image and answer the question: Does the highlighted object meet the query description? Your answer should be "Yes" or "No".\\
    
Your answer:
\end{prompt}

\end{document}